\documentclass{esannV2}
\usepackage[dvips]{graphicx}
\usepackage[latin1]{inputenc}
\usepackage{amssymb,amsmath,array}
\usepackage{stmaryrd} % \lightning

\usepackage{xcolor}
\usepackage{tikz}
\usetikzlibrary{shapes.geometric, arrows, fit, decorations.pathreplacing, calc, positioning}
\usepackage{natbib}
\newcommand{\model}{\Phi}
\newcommand{\R}{\mathbb{R}}
\newcommand{\outputspace}{\mathcal{Y}}
\newcommand{\globalexplanation}{E}
\newcommand{\localexplainer}{L_\model}

\newcommand{\instanceset}{X}
\newcommand{\instance}{\mathbf{x}}

\newcommand{\class}{c}

\newcommand{\epruned}{E _{\text{p}}}
\newcommand{\ecand}{E_{p}'}

\newcommand{\pruningalg}{{\sc ThresholdPruning}}
\newcommand{\cfire}{{CFIRE}}
\newcommand{\longalgname}{{\sc Closed Frequent Itemset Rules from Explanations}}

\usepackage{booktabs}
\usepackage{algorithmicx}
\usepackage{algorithm}
\usepackage[noend]{algpseudocode}
\usepackage{placeins}
\usepackage{hyperref}
\begin{document}
%style file for ESANN manuscripts
% \title{Improving Compactness and Ambiguity of CFIRE Rule Explanations}
% \title{Pruning CFIRE Rule Explanations for Compactness and Lower Ambiguity}
\title{Improving Compactness and Reducing Ambiguity of CFIRE Rule-Based Explanations}

%***********************************************************************
% AUTHORS INFORMATION AREA
%***********************************************************************
\author{Sebastian M\"uller$^{1,2}$, 
Tobias Schneider$^{1}$, 
Ruben Kemna$^{1}$ 
and Vanessa Toborek$^{1,2}$
%
% Optional short acknowledgment: remove next line if non-needed
% \thanks{This is an optional funding source acknowledgement.}
%
% DO NOT MODIFY THE FOLLOWING '\vspace' ARGUMENT
\vspace{.3cm}\\
%
% Addresses and institutions (remove "1- " in case of a single institution)
% $^1$ Machine Learning and Artificial Intelligence Group, University of Bonn, Germany 
% \vspace{.1cm}\\
$^1$ University of Bonn, Bonn, Germany
\vspace{.1cm}\\
$^2$ Lamarr Institute, Bonn, Germany
}
%***********************************************************************
% END OF AUTHORS INFORMATION AREA
%***********************************************************************

\maketitle

\begin{abstract}
Models trained on tabular data are widely used in sensitive domains, increasing the demand for explanation methods to meet transparency needs.
CFIRE is a recent algorithm in this domain that constructs compact surrogate rule models from local explanations. 
While effective, CFIRE may assign rules associated with different classes to the same sample, introducing ambiguity.
We investigate this ambiguity and propose a post-hoc pruning strategy that removes rules with low contribution or conflicting coverage, yielding smaller and less ambiguous models while preserving fidelity. 
Experiments across multiple datasets confirm these improvements with minimal impact on predictive performance.
\end{abstract}

\section{Introduction}
\label{sec:intro}
Machine learning models for tabular data are widely used in domains such as finance or healthcare, and cybersecurity
~\cite{borisov2024dnntabularsurvey}. These applications involve sensitive information and often affect individuals directly. As a result, there is an increased need for transparency, making model explanations an important component of such systems.
Rule-based explanations are a natural fit for tabular data that provide compact, human-readable descriptions~\cite{ribeiro2018anchors,mersha2024xaisurvey}. Their usefulness, however, depends not only on fidelity but also on properties such as size and the absence of conflicting or ambiguous statements~\cite{lakkaraju2016interpretable,nauta2023anecdotal}.
% Closed Frequent Itemset Rules from Explanations
\longalgname{}~(\cfire{})~\cite{muller2025cfire} is a previously proposed algorithm that constructs global rule models by aggregating local feature-importance explanations. \cfire{} is model-agnostic and can be used with any local explainer that identifies important dimensions for individual predictions. The algorithm produces disjunctive normal form (DNF) rule models that are compact and have competitive fidelity compared to existing rule-extraction methods.
One observation regarding the \cfire{} algorithm is that in some cases multiple rules, \textit{associated with different classes}, are applicable to the same datapoint. 
To obtain a single label prediction, \cfire{} chooses the rule with the highest accuracy on the training set. 
While this yields correct predictions, the presence of several applicable rules introduces ambiguity in the resulting explanation.

In this work, we study the inter-class ambiguity of \cfire{} rule models and propose a post-hoc pruning strategy that removes rules that contribute little to predictive performance or primarily generate ambiguous coverage. 
We test our pruning-strategy on 2\,100 \cfire{} rule models.
The resulting rule models are more compact and show reduced ambiguity while preserving fidelity.
Code for the pruning extension available at \href{https://github.com/semueller/CFIRE/tree/pruning}{github.com/semueller/CFIRE/tree/pruning}.

%***********************************************************************
%***********************************************************************

\section{Background: \cfire{}}
\label{sec:cfire}
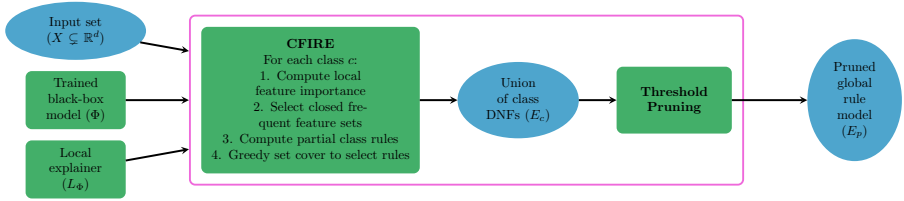
\begin{figure}[t]
  \centering
  \label{fig:cfire}
  \resizebox{\textwidth}{!}{
    \tikzstyle{artifact} = [ellipse, minimum height = 15mm, text width=30mm, text centered, line width = 0.5mm, draw={rgb,255:red,78; green,165; blue,205}, fill={rgb,255:red,78; green,165; blue,205}, fill opacity=0.3, text opacity=1]

\tikzstyle{method} = [rectangle, draw, text centered, draw={rgb,255:red,67; green,175; blue,105}, line width = 0.5mm, fill={rgb,255:red,67; green,175; blue,105}, fill opacity=0.3, text opacity=1, minimum height = 15mm, rounded corners, text width=50mm]

\newcommand{\arrowwidth}[0]{1.5pt}
\tikzstyle{arrow} = [thick, ->, >=stealth, line width=\arrowwidth]

\begin{tikzpicture}[node distance=20mm]

        % Main input nodes
        \node(data)[artifact, inner xsep=3mm, text width=20mm]{Input set ($\instanceset \subsetneq \R^d$)};
        \node(blackbox)[method, inner xsep=3mm, text width=20mm, below = 3mm of data]{Trained black-box model ($\model$)};
        \node(explain)[method, inner xsep=3mm, text width=20mm, below = 3mm of blackbox]{Local explainer ($\localexplainer$)};

        % CFIRE internal components horizontally
        \node[method, inner ysep=3mm, text width=55mm, right=20mm of blackbox] (cfire) {\textbf{CFIRE}\\For each class $c$:\\1. Compute local feature importance\\2. Select closed frequent feature sets \\3. Compute partial class rules \\4. Greedy set cover to select rules};
        \node[artifact, text width = 20mm, right=10mm of cfire] (dnf) {Union of class DNFs ($\globalexplanation_c$)};
        \node[method, inner ysep=5mm, inner xsep=5mm, text width=20mm, right=10mm of dnf] (prune) {\textbf{Threshold Pruning}};
        \node[artifact, text width = 15mm, right=20mm of prune] (global) {Pruned global rule model ($\globalexplanation_p$)};

        % Outer box to mark our algorithm block
        \node[inner ysep=3mm, inner xsep=3mm, fit=(cfire)(dnf)(prune), draw={rgb,255:red,239; green,105; blue,219}, line width = 0.5mm, rounded corners, text height=20mm, align=center] (us) {};

        % All simple horizontal arrows
        \draw[arrow] (data) -- (us);
        \draw[arrow] (blackbox) -- (us);
        \draw[arrow] (explain) -- (us);
        \draw[arrow] (cfire) -- (dnf);
        \draw[arrow] (dnf) -- (prune);
        \draw[arrow] (prune) -- (global);

        % Loop-arrow for iteration over classes
        % Intermediate points of the arrow
        % \coordinate (intermediate1) at ($(select.south)+(0, -12mm)$);
        % \coordinate (intermediate2) at ($(attributes.south)+(0, -12mm)$);

        % Vertical segment from 'select' to the first intermediate point
        % \draw[-, thick, line width = \arrowwidth] (select.south) -- (intermediate1);
        % Horizontal segment with label
        % \draw[-, thick, line width = \arrowwidth] (intermediate1) -- node[midway, below]{for each class} (intermediate2);
        % Final vertical segment
        % \draw[arrow] (intermediate2) -- (attributes.south);

\end{tikzpicture}
  }
  \caption{Schematic overview of the original \cfire{} pipeline with the proposed \pruningalg{} (Algorithm~\ref{alg:pruning}) extension.% Given black-box model $\model$, local explainer $\localexplainer$, and input data $\instanceset$, our algorithm computes DNFs as explanations for each class independently.
  }
\end{figure}
\cfire{} constructs global rule models by aggregating local feature-importance explanations for a trained black-box classifier $\model\!:\R^d\to\outputspace$.
Let $\instanceset \subset \R^d$ denote the samples used for rule extraction.
For each class $\class \in \outputspace$, \cfire{} identifies important dimensions for instances in
$X_\class = \{\instance \in \instanceset : \model(\instance) = \class\}$ using a local explainer~$\localexplainer$. These important dimensions are treated as transactions, and closed frequent itemsets are mined to obtain candidate terms.
Each itemset is converted into an axis-aligned box constraint describing a candidate literal conjunction.
Still independently for each class, a compact DNF is obtained by selecting a subset of these candidate terms via a greedy set-cover procedure. This yields a class-conditional DNF $E_\class$. The procedure is repeated for all classes.
Each $E_\class$ is a finite set of \emph{terms}, where a term $t \in E_\class$ is a conjunction of interval constraints and predicts class $\class$.  We write $\class(t)=\class$ to denote the class associated with $t$.
The global rule model is the union $\globalexplanation = \bigcup_{\class \in \outputspace} E_\class$.
A term is \emph{applicable} to a sample $\instance$ if all its constraints are satisfied, written $t(\instance)=1$.

Since the per-class DNFs are constructed independently, multiple rules from different classes may apply to the same sample. \cfire{} resolves such cases by a tie-breaking mechanism: among all applicable rules, the one with highest empirical accuracy on~$\instanceset$ is selected. While preserving fidelity, it leaves the explanation ambiguous whenever several class rules are simultaneously applicable. Figure~\ref{fig:cfire} shows the \cfire{} pipeline extended with the pruning step we introduce.

\textbf{Inter-class ambiguity.} For a rule model $\globalexplanation = \bigcup_{\class\in\outputspace} E_\class$, let $\mathrm{App}(\instance) = \{\,t \in \globalexplanation : t(\instance)=1\,\}$ denote the set of all rules applicable to a sample~$\instance$.
We focus on \emph{inter-class} ambiguity, i.e., cases where applicable rules originate from more than one class. A sample $\instance$ is ambiguous if $\{\class(t) : t \in \mathrm{App}(\instance)\}$ contains more than one class.
Given a reference set $\instanceset$, the ambiguity on $\globalexplanation$ is defined as
% \vspace{-0.5ex} 

\smallskip \qquad \quad
$
\mathrm{Amb}(E;X) \;=\;
\frac{1}{|X|}\,\bigl|\{\,\instance \in X :
|\{\class(t) : t\in\mathrm{App}(\instance)\}| > 1 \,\}\bigr|
$
% \vspace{-0.5ex}

\smallskip
% Because class-wise DNFs in \cfire{} are constructed independently, ambiguity may occur even when predictive fidelity is high.
We use this measure to report the inter-class ambiguity of original (unpruned) \cfire{} models on a test set together with their F1-scores and model sizes (total number of terms) in Table~\ref{tab:original}.
These results show that inter-class ambiguity can be substantial across datasets and local explainers, even when predictive fidelity of the rules is very high. 
This motivates extending the CFIRE pipeline by our pruning strategy we will now describe in the next section.

\begin{table}
    \centering
    \resizebox{0.80\textwidth}{!}{%
        % task abalone		 -> mean acc 0.64
% task breastcancer		 -> mean acc 1.00
% task ionosphere		 -> mean acc 0.94
% task btsc		 -> mean acc 0.80
% task spf		 -> mean acc 0.75
% task spambase		 -> mean acc 0.93
% task heloc		 -> mean acc 0.73
% task autouniv		 -> mean acc 0.41
% task wine		 -> mean acc 0.99
% task iris		 -> mean acc 0.92
% task vehicle		 -> mean acc 0.79
% task diggle		 -> mean acc 0.95
% task beans		 -> mean acc 0.91
% task breastw		 -> mean acc 0.99
\begin{tabular}{l|lccc}
\toprule
Dataset & Metric & CFIRE-KS & CFIRE-LI & CFIRE-IG \\
\midrule
abalone & F1 & 0.81{\scriptsize$\pm$0.02} & 0.80{\scriptsize$\pm$0.03} & 0.75{\scriptsize$\pm$0.02} \\
$|X|=626$ & Size & 31.3{\scriptsize$\pm$6.4} & 21.7{\scriptsize$\pm$5.5} & 19.3{\scriptsize$\pm$5.6} \\
mean accuracy $\Phi$=0.64 & Amb & 32.7{\scriptsize$\pm$4.6} & 23.9{\scriptsize$\pm$4.7} & 21.7{\scriptsize$\pm$5.4} \\
\midrule
autouniv & F1 & 0.43{\scriptsize$\pm$0.04} & 0.41{\scriptsize$\pm$0.04} & 0.44{\scriptsize$\pm$0.04} \\
$|X|=140$ & Size & 22.7{\scriptsize$\pm$5.1} & 24.4{\scriptsize$\pm$4.9} & 20.1{\scriptsize$\pm$4.4} \\
mean accuracy $\Phi$=0.41 & Amb & 58.5{\scriptsize$\pm$13.1} & 45.8{\scriptsize$\pm$12.3} & 37.8{\scriptsize$\pm$17.3} \\
\midrule
beans & F1 & 0.92{\scriptsize$\pm$0.01} & 0.89{\scriptsize$\pm$0.03} & 0.92{\scriptsize$\pm$0.01} \\
$|X|=1050$ & Size & 37.1{\scriptsize$\pm$6.3} & 36.3{\scriptsize$\pm$6.4} & 27.2{\scriptsize$\pm$4.3} \\
mean accuracy $\Phi$=0.91 & Amb & 35.6{\scriptsize$\pm$9.3} & 42.1{\scriptsize$\pm$13.5} & 29.3{\scriptsize$\pm$9.3} \\
\midrule
breastcancer & F1 & 0.89{\scriptsize$\pm$0.04} & 0.82{\scriptsize$\pm$0.07} & 0.88{\scriptsize$\pm$0.03} \\
$|X|=170$ & Size & 6.9{\scriptsize$\pm$2.1} & 5.6{\scriptsize$\pm$2.1} & 2.6{\scriptsize$\pm$1.0} \\
mean accuracy $\Phi$=0.99 & Amb & 18.2{\scriptsize$\pm$5.5} & 12.6{\scriptsize$\pm$6.5} & 24.6{\scriptsize$\pm$6.6} \\
\midrule
breastw & F1 & 0.97{\scriptsize$\pm$0.01} & 0.95{\scriptsize$\pm$0.02} & 0.91{\scriptsize$\pm$0.03} \\
$|X|=1000$ & Size & 15.8{\scriptsize$\pm$4.3} & 10.2{\scriptsize$\pm$3.4} & 2.9{\scriptsize$\pm$1.0} \\
mean accuracy $\Phi$=0.99 & Amb & 19.6{\scriptsize$\pm$3.3} & 16.9{\scriptsize$\pm$4.8} & 14.7{\scriptsize$\pm$10.6} \\
\midrule
btsc & F1 & 0.92{\scriptsize$\pm$0.02} & 0.91{\scriptsize$\pm$0.02} & 0.90{\scriptsize$\pm$0.01} \\
$|X|=150$ & Size & 3.8{\scriptsize$\pm$1.1} & 3.5{\scriptsize$\pm$1.8} & 2.4{\scriptsize$\pm$1.3} \\
mean accuracy $\Phi$=0.80 & Amb & 13.0{\scriptsize$\pm$6.9} & 21.8{\scriptsize$\pm$2.9} & 3.3{\scriptsize$\pm$4.7} \\
\midrule
diggle & F1 & 0.85{\scriptsize$\pm$0.02} & 0.84{\scriptsize$\pm$0.02} & 0.69{\scriptsize$\pm$0.05} \\
$|X|=62$ & Size & 9.6{\scriptsize$\pm$0.6} & 10.4{\scriptsize$\pm$0.7} & 13.0{\scriptsize$\pm$1.4} \\
mean accuracy $\Phi$=0.95 & Amb & 2.3{\scriptsize$\pm$1.6} & 6.2{\scriptsize$\pm$18.5} & 5.7{\scriptsize$\pm$11.3} \\
\midrule
heloc & F1 & 0.87{\scriptsize$\pm$0.03} & 0.87{\scriptsize$\pm$0.01} & 0.85{\scriptsize$\pm$0.02} \\
$|X|=1974$ & Size & 23.3{\scriptsize$\pm$7.4} & 8.4{\scriptsize$\pm$5.3} & 10.5{\scriptsize$\pm$5.5} \\
mean accuracy $\Phi$=0.73 & Amb & 79.3{\scriptsize$\pm$7.3} & 30.2{\scriptsize$\pm$15.6} & 29.4{\scriptsize$\pm$14.6} \\
\midrule
ionosphere & F1 & 0.74{\scriptsize$\pm$0.06} & 0.86{\scriptsize$\pm$0.03} & 0.74{\scriptsize$\pm$0.05} \\
$|X|=104$ & Size & 2.4{\scriptsize$\pm$0.9} & 4.4{\scriptsize$\pm$1.0} & 4.6{\scriptsize$\pm$1.5} \\
mean accuracy $\Phi$=0.94 & Amb & 88.4{\scriptsize$\pm$22.6} & 32.1{\scriptsize$\pm$10.1} & 15.2{\scriptsize$\pm$4.8} \\
\midrule
iris & F1 & 0.87{\scriptsize$\pm$0.02} & 0.87{\scriptsize$\pm$0.03} & 0.83{\scriptsize$\pm$0.01} \\
$|X|=37$ & Size & 4.7{\scriptsize$\pm$0.7} & 3.1{\scriptsize$\pm$0.2} & 5.9{\scriptsize$\pm$0.3} \\
mean accuracy $\Phi$=0.92 & Amb & 14.6{\scriptsize$\pm$3.6} & 9.9{\scriptsize$\pm$5.0} & 20.9{\scriptsize$\pm$2.8} \\
\midrule
spambase & F1 & 0.83{\scriptsize$\pm$0.07} & 0.83{\scriptsize$\pm$0.01} & 0.87{\scriptsize$\pm$0.01} \\
$|X|=921$ & Size & 15.9{\scriptsize$\pm$7.3} & 8.8{\scriptsize$\pm$3.1} & 4.9{\scriptsize$\pm$1.3} \\
mean accuracy $\Phi$=0.93 & Amb & 49.0{\scriptsize$\pm$18.7} & 6.8{\scriptsize$\pm$4.0} & 34.7{\scriptsize$\pm$3.2} \\
\midrule
spf & F1 & 0.74{\scriptsize$\pm$0.03} & 0.73{\scriptsize$\pm$0.03} & 0.76{\scriptsize$\pm$0.03} \\
$|X|=389$ & Size & 18.1{\scriptsize$\pm$5.6} & 14.5{\scriptsize$\pm$4.9} & 14.0{\scriptsize$\pm$5.5} \\
mean accuracy $\Phi$=0.75 & Amb & 56.5{\scriptsize$\pm$12.8} & 48.5{\scriptsize$\pm$14.6} & 49.3{\scriptsize$\pm$17.0} \\
\midrule
vehicle & F1 & 0.53{\scriptsize$\pm$0.04} & 0.52{\scriptsize$\pm$0.04} & 0.56{\scriptsize$\pm$0.04} \\
$|X|=170$ & Size & 22.1{\scriptsize$\pm$4.5} & 20.9{\scriptsize$\pm$4.0} & 17.3{\scriptsize$\pm$4.0} \\
mean accuracy $\Phi$=0.79 & Amb & 46.4{\scriptsize$\pm$18.7} & 29.0{\scriptsize$\pm$10.7} & 30.8{\scriptsize$\pm$10.0} \\
\midrule
wine & F1 & 0.66{\scriptsize$\pm$0.06} & 0.70{\scriptsize$\pm$0.07} & 0.39{\scriptsize$\pm$0.04} \\
$|X|=36$ & Size & 4.1{\scriptsize$\pm$1.1} & 5.2{\scriptsize$\pm$1.1} & 5.3{\scriptsize$\pm$1.5} \\
mean accuracy $\Phi$=0.99 & Amb & 7.2{\scriptsize$\pm$4.6} & 1.9{\scriptsize$\pm$3.1} & 9.9{\scriptsize$\pm$9.0} \\
\bottomrule
\end{tabular}

    }
    \caption{$|X|$: \#rule-extraction samples; mean accuracy over the 50 black-boxes trained per task. Baseline F1, rule count (Size), and inter-class ambiguity (Amb) of original \cfire{} models for local explainers KernelShap (KS), LIME (LI) and Integrated Gradients (IG).}
    \label{tab:original}
\end{table}
% for easier access these tables contain results for safe and best pruning:
% \begin{table}
%     \centering
%     \resizebox{0.67\textwidth}{!}{%
%         \input{tables/overview_safe}
%     }
%     \caption{SAFE}
%     \label{tab:overview}
% \end{table}\begin{table}
%     \centering
%     \resizebox{0.67\textwidth}{!}{%
%         \input{tables/overview_best}
%     }
%     \caption{BEST}
%     \label{tab:overview}
% \end{table}
% \newpage

\begin{algorithm}
\caption{\sc ThresholdPruning}
\label{alg:pruning}
\textbf{Input:} Rule model $\globalexplanation$, samples $\instanceset$, labels $Y$, performance tolerance $\theta \in [0,1]$ \\
\textbf{Output}: Pruned rule model $\epruned$
\begin{algorithmic}[1]
    \State $\epruned \gets \globalexplanation$
    \State $\text{Wins} \gets \{ \text{t}\,:\, 0 \text{ for t} \in \globalexplanation \}$  \Comment{Map to associate each term with an integer}
    \State $\text{A} \gets \{ t\,:\,$\Call{Accuracy}{$t, \instanceset, Y$} for $ t \in \globalexplanation\}$ \Comment{Compute accuracy for each term}
    \ForAll{$\instance \in \instanceset$}
        \State $T_x \gets$ \Call{GetAllApplicableTerms}{$\globalexplanation, \instance$}
        % \State $t_\instance \gets$ \Call{ChooseSingleMostAccurate}{A, $T_\instance$}
        \State $t_\instance \gets$ $\underset{t \in T_\instance}{\arg \max}$ $\text{A}[t]$ \Comment{Tie-breaking as in CFIRE}
        \State Wins$[t_\instance]$ += 1
    \EndFor
    \State $\ecand \gets \epruned,\ k \gets 0$
    \While{ \Call{Accuracy}{$\ecand, X, Y$} $>$ $(1-\theta)$\Call{Accuracy}{$\globalexplanation, \instanceset, Y$}}
        \State $\epruned \gets \ecand$
        \State $\ecand \gets \epruned \setminus \{t\, |\, t\in \epruned: \text{Wins}[t]<=k\}$  %\Comment{Remove all terms with $<=k$ wins}
        \State $k \gets k+1$
    \EndWhile
    \State \Return $\epruned$
\end{algorithmic}

\end{algorithm}

\section{Rule Pruning}
\label{sec:pruning}

Ambiguity analysis shows that \cfire{} rule models may contain terms that rarely affect predictions, yet contribute to inter-class ambiguity. We therefore introduce a post-hoc pruning procedure that removes such terms while preserving the model's predictive behaviour.

For each term $t$ in the global explanation $\globalexplanation$, we compute its \textit{win-count}, denoted by $\mathrm{Wins}[t]$, on a reference set $X$: the number of instances where $t$ is selected by the \cfire{} tie-breaking mechanism.

\hspace*{-0.3cm}\textbf{Threshold pruning:}
Given a rule model $\globalexplanation$, labeled reference data $(\instanceset, \Phi(\instanceset))$ and a tolerance threshold $\theta\in[0,1]$, we iteratively remove terms in ascending order of $\mathrm{Wins}[t]$ as long as the model's relative accuracy change on $X$ remains below $\theta$.
% while the relative change in accuracy of the resulting model on $X$ is no more than $\theta$. %This yields a pruned model $E^{\theta}$ that balances compactness and fidelity.
Setting $\theta=0$ yields the special case of \textbf{safe pruning}, removing only terms with zero wins and thus preserving predictive behaviour on $X$.
Algorithm~\ref{alg:pruning} summarises the \pruningalg{} procedure.

\section{Experiments}
\label{sec:experiments}
We apply \pruningalg{} to the original \cfire{} rule models, with $\theta\in\{0.0, 0.05\}$ using $\instanceset$ for both a safe and moderate pruning variant. 
% Following the experimental protocol of the original \cfire{} paper, we use the same 14 datasets, data splits, and black-box neural networks. 
The \cfire{} models were computed for 700 small neural networks, trained on 14 different tasks. \cfire{} was applied to their predictions on the reference set $X$ using three different local explainers: KernelShap (KS), LIME (LI), and Integrated Gradients (IG). 
% All \cfire{} hyperparameters are set to default. 
We evaluate the original and pruned rule models on a held-out set $\instanceset_\text{test}$. 
We report the average relative changes after pruning for Size, F1, and inter-class ambiguity (in \%) over the original metric scores shown in Table~\ref{tab:original}.
% . All rule models are evaluated on the original test sets reporting F1 and inter-class ambiguity. % as defined.% in Section~\ref{sec:cfire}. 
% In Table~\ref{tab:pruning}, w
% We report relative changes in \% over the original metric scores shown in Table~\ref{tab:original}.

\newpage
\begin{table}[h]
    \centering
    \begin{tabular}{llccc}
\toprule
$\theta$ & Metric ($\Delta\%$) & CFIRE-KS & CFIRE-LI & CFIRE-IG \\
\midrule
0.0 & $F1_{\instanceset}$ & 0.00{\scriptsize$\pm$0.00} & 0.00{\scriptsize$\pm$0.00} & 0.00{\scriptsize$\pm$0.00} \\
 & $F1_{}$ & -0.01{\scriptsize$\pm$0.38} & -0.01{\scriptsize$\pm$0.37} & 0.05{\scriptsize$\pm$0.64} \\
 & $\mathrm{Size}$ & -15.51{\scriptsize$\pm$17.30} & -7.63{\scriptsize$\pm$12.06} & -4.63{\scriptsize$\pm$9.41} \\
 % & $\text{Amb}_{\instanceset}$ & -20.43{\scriptsize$\pm$31.11} & -9.16{\scriptsize$\pm$21.66} & -6.36{\scriptsize$\pm$16.94} \\
 & $\text{Amb}_{}$ & -20.64{\scriptsize$\pm$31.31} & -9.02{\scriptsize$\pm$21.55} & -5.43{\scriptsize$\pm$15.36} \\
 \midrule
0.05 & $F1_{\instanceset}$ & -1.46{\scriptsize$\pm$1.73} & -1.58{\scriptsize$\pm$1.36} & -1.36{\scriptsize$\pm$1.22} \\
 & $F1_{}$ & -0.04{\scriptsize$\pm$2.11} & -0.06{\scriptsize$\pm$1.57} & -0.09{\scriptsize$\pm$1.79} \\
 & $\mathrm{Size}$ & -45.43{\scriptsize$\pm$27.72} & -36.58{\scriptsize$\pm$24.69} & -31.12{\scriptsize$\pm$24.39} \\
 % & $\text{Amb}_{\instanceset}$ & -46.27{\scriptsize$\pm$38.01} & -33.46{\scriptsize$\pm$35.40} & -30.62{\scriptsize$\pm$34.50} \\
 & $\text{Amb}_{}$ & -47.01{\scriptsize$\pm$38.08} & -31.68{\scriptsize$\pm$34.92} & -32.33{\scriptsize$\pm$35.69} \\
\bottomrule
\end{tabular}

% \begin{tabular}{llccc}
% \toprule
% Strat & Metric ($\Delta\%$) & IG & KS & LI \\
% \midrule
% Safe & $F1_{\mathrm{val}}$ & 0.00{\scriptsize$\pm$0.00} & 0.00{\scriptsize$\pm$0.00} & 0.00{\scriptsize$\pm$0.00} \\
%  & $F1_{\mathrm{test}}$ & -0.04{\scriptsize$\pm$0.86} & -0.03{\scriptsize$\pm$0.95} & -0.09{\scriptsize$\pm$0.89} \\
%  & $\mathrm{Size}$ & -5.35{\scriptsize$\pm$9.44} & -17.74{\scriptsize$\pm$17.38} & -8.46{\scriptsize$\pm$13.34} \\
%  & $\text{Amb}_{\text{val}}$ & -7.86{\scriptsize$\pm$19.07} & -30.17{\scriptsize$\pm$35.89} & -15.56{\scriptsize$\pm$29.10} \\
%  & $\text{Amb}_{\text{test}}$ & -7.50{\scriptsize$\pm$18.93} & -30.60{\scriptsize$\pm$36.07} & -16.65{\scriptsize$\pm$29.54} \\
%  \midrule
% Best & $F1_{\mathrm{val}}$ & -1.91{\scriptsize$\pm$1.68} & -2.33{\scriptsize$\pm$2.41} & -2.02{\scriptsize$\pm$2.34} \\
%  & $F1_{\mathrm{test}}$ & -0.66{\scriptsize$\pm$2.13} & -0.77{\scriptsize$\pm$2.82} & -0.79{\scriptsize$\pm$2.94} \\
%  & $\mathrm{Size}$ & -33.89{\scriptsize$\pm$24.37} & -44.81{\scriptsize$\pm$28.50} & -35.15{\scriptsize$\pm$28.48} \\
%  & $\text{Amb}_{\text{val}}$ & -28.16{\scriptsize$\pm$31.82} & -52.93{\scriptsize$\pm$36.98} & -37.31{\scriptsize$\pm$37.96} \\
%  & $\text{Amb}_{\text{test}}$ & -26.50{\scriptsize$\pm$29.28} & -55.63{\scriptsize$\pm$37.30} & -40.66{\scriptsize$\pm$38.82} \\
% \bottomrule
% \end{tabular}

    \caption{Mean$\pm$std of \textit{relative} changes in F1, Size, and Ambiguity after pruning.}
    \label{tab:pruning}
\end{table}
\begin{figure}[h!]
    \centering
    \includegraphics[width=\linewidth]{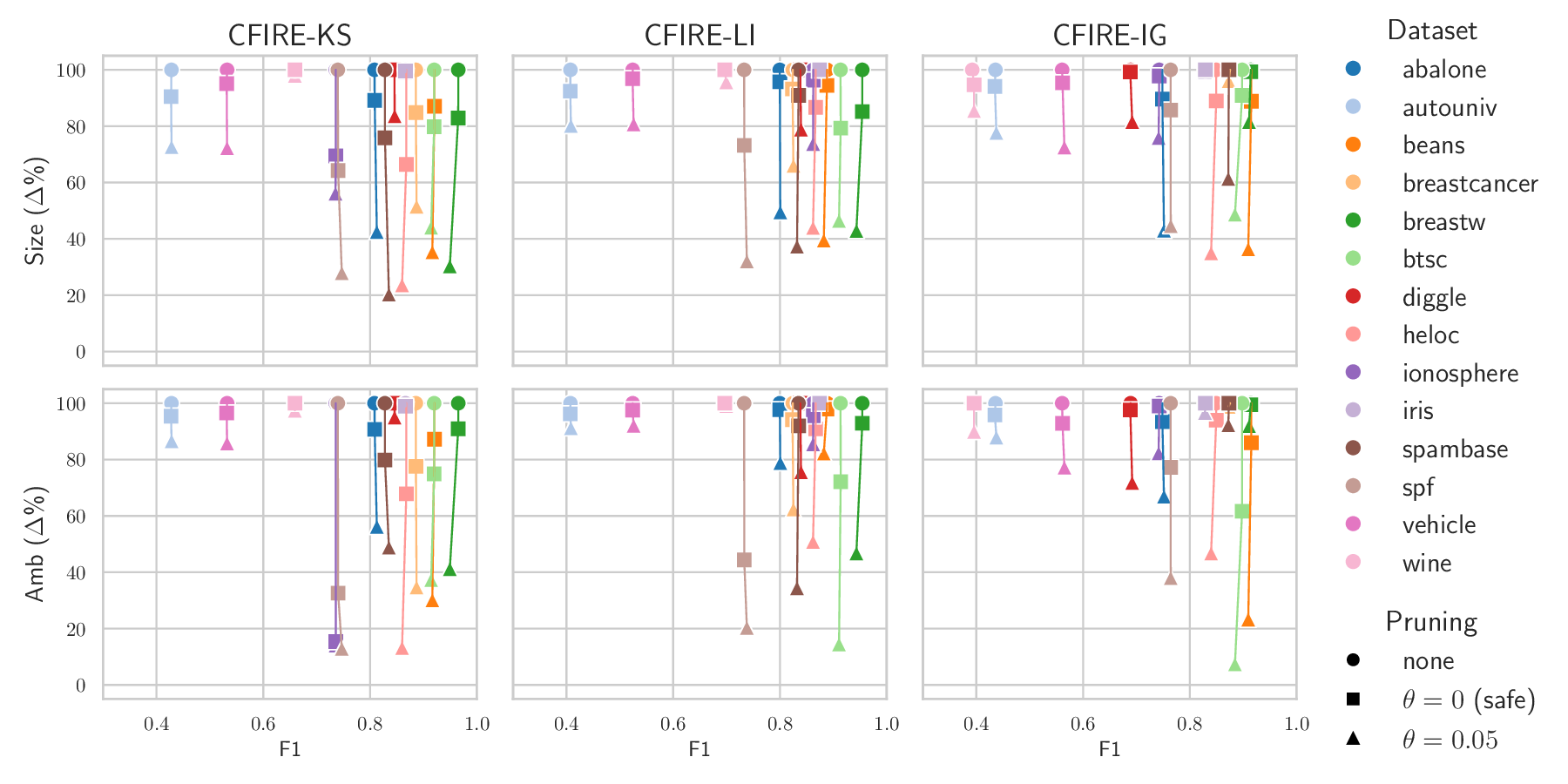}
    \caption{Trade-off between F1 and relative effects of pruning on Size and Ambiguity. Color encodes dataset; markers circle, square and triangle show means over $\Phi$s for no pruning, safe pruning $\theta=0$ and threshold pruning $\theta=0.05$.}
    \label{fig:pruning}
\end{figure}

\paragraph{\textbf{Results}} Table~\ref{tab:pruning} provides an aggregate overview across all datasets and black-boxes. As expected, safe pruning $\theta=0$ leaves the F1 score on $X$ unchanged. Surprisingly, this already yields noticeable reductions in size and ambiguity, indicating that a nontrivial fraction of terms is never decisive for prediction. Even for $\theta=0.05$ the average performance loss remains below $2\%$, with standard deviations under $3\%$ in all settings. In contrast, the reductions in model size and inter-class ambiguity are large. Notably, the final performance remains well above the $5\%$ tolerance, suggesting that \cfire{} rule models often contain many low coverage or redundant terms.

Figure~\ref{fig:pruning} shows the changes per dataset. The x-axis displays F1 scores on the test set, while the y-axis shows the relative change in Size (upper row) and Amb (lower row). Circles denote the original models, squares safe pruning, and triangles Threshold pruning; markers placed at mean values for all $\Phi$.

Again we see, improvements come at virtually no cost to predictive performance across all datasets. Strong and consistent gains across all local explainers are seen for \textsc{heloc}, \textsc{spf}, and, perhaps surprisingly, \textsc{btsc}, where ambiguity often drops by more than half and model sizes shrink substantially. Several datasets show notable improvements for at least one explainer, such as \textsc{ionosphere} (KS), \textsc{spambase} (KS/LI), and \textsc{beans} (KS/IG). In contrast, very small datasets such as \textsc{wine}, \textsc{iris}, or \textsc{diggle} change little, and in some cases ambiguity barely moves even though size decreases.
Larger \cfire{} models generally benefit more from pruning. 
Datasets such as \textsc{abalone}, \textsc{vehicle}, \textsc{beans}, and \textsc{autouniv} yield the largest \cfire{} models, often with more than 20--30 terms before pruning, and these are also the tasks where ambiguity is highest to begin with.
While pruning reduces size and sometimes ambiguity, several of these tasks remain highly ambiguous. In particular, \textsc{autouniv} shows persistent ambiguity ($30$--$50\%$) despite reductions in rule count, indicating that some tasks induce inherently overlapping rule regions.

\paragraph{\textbf{Conclusion}} Our study shows that a post-hoc rule-pruning step can address a key limitation of \cfire{} models. Overall, pruning reliably improves compactness and reduces ambiguity at close to no cost to accuracy in many cases. It is effective across datasets and explainers, although some tasks remain intrinsically ambiguous.
\smallskip

% ****************************************************************************
% BIBLIOGRAPHY AREA
% ****************************************************************************

\begin{footnotesize}

% IF YOU DO NOT USE BIBTEX, USE THE FOLLOWING SAMPLE SCHEME FOR THE REFERENCES
% ----------------------------------------------------------------------------
% ----------------------------------------------------------------------------

% IF YOU USE BIBTEX,
% - DELETE THE TEXT BETWEEN THE TWO ABOVE DASHED LINES
% - UNCOMMENT THE NEXT TWO LINES AND REPLACE 'Name_Of_Your_BibFile'

\bibliographystyle{unsrt}
\bibliography{bib_seb}

@inproceedings{muller2025cfire,
  title={{CFIRE: A General Method for Combining Local Explanations}},
  author={M{\"u}ller, Sebastian and Toborek, Vanessa and Horv{\'a}th, Tam{\'a}s and Bauckhage, Christian},
  booktitle={``World Conference on Explainable Artificial Intelligence'' Conference Series in Communications in Computer and Information Science},
  year={2025},
  journal={CCIS},
  volume={2557},
  organization={Springer}
}

@inproceedings{ribeiro2018anchors,
  title={{Anchors: High-Precision Model-Agnostic Explanations}},
  author={Ribeiro, Marco Tulio and Singh, Sameer and Guestrin, Carlos},
  booktitle={{Proceedings of the AAAI Conference on Artificial Intelligence}},
  volume= {32},
  number={1},
  year={2018}
}

@inproceedings{lakkaraju2016interpretable,
  title={{Interpretable decision sets: A joint framework for description and prediction}},
  author={Lakkaraju, Himabindu and Bach, Stephen H and Leskovec, Jure},
  booktitle={Proceedings of the 22nd ACM SIGKDD international conference on knowledge discovery and data mining},
  year={2016},
series = {KDD '16},
publisher = {Association for Computing Machinery},
}

@article{nauta2023anecdotal,
  title={{From Anecdotal Evidence to Quantitative Evaluation Methods: A Systematic Review on Evaluating Explainable AI}},
  author={Nauta, Meike and Trienes, Jan and Pathak, Shreyasi and Nguyen, Elisa and Peters, Michelle and Schmitt, Yasmin and Schl{\"o}tterer, J{\"o}rg and Van Keulen, Maurice and Seifert, Christin},
  journal={ACM Computing Surveys},
  volume={volume 55},
  year={2023},
  publisher = {Association for Computing Machinery}
}

@article{mersha2024xaisurvey,
  title = {{Explainable Artificial Intelligence: {{A}} Survey of Needs, Techniques, Applications, and Future Direction}},
  shorttitle = {Explainable Artificial Intelligence},
  author = {Mersha, Melkamu and Lam, Khang and Wood, Joseph and AlShami, Ali K. and Kalita, Jugal},
  year = {2024},
  journal = {Neurocomputing},
  volume = {volume 599},
  issn = {09252312},
  doi = {10.1016/j.neucom.2024.128111},
  url = {https://linkinghub.elsevier.com/retrieve/pii/S0925231224008828},
}

@article{borisov2024dnntabularsurvey,
	title        = {{Deep {{Neural Networks}} and {{Tabular Data}}: {{A Survey}}}},
	shorttitle   = {Deep {{Neural Networks}} and {{Tabular Data}}},
	author       = {Borisov, Vadim and Leemann, Tobias and Se{\ss}ler, Kathrin and Haug, Johannes and Pawelczyk, Martin and Kasneci, Gjergji},
	volume       = {volume 35},
	doi          = {10.1109/TNNLS.2022.3229161},
	issn         = {2162-2388},
	url          = {https://ieeexplore.ieee.org/document/9998482/},
	urldate      = {2025-11-19},
	year         = {2024},
	journal = {IEEE Transactions on Neural Networks and Learning Systems}
}

\end{footnotesize}

% ****************************************************************************
% END OF BIBLIOGRAPHY AREA
% ****************************************************************************

\end{document}